\documentclass[10pt,twocolumn,letterpaper]{article}

\usepackage[pagenumbers]{cvpr} 

\usepackage{graphicx}
\usepackage{amsmath}
\usepackage{amssymb}
\usepackage{booktabs}
\usepackage{multirow}
\usepackage[accsupp]{axessibility}  

\usepackage[pagebackref,breaklinks,colorlinks]{hyperref}

\usepackage[capitalize]{cleveref}
\crefname{section}{Sec.}{Secs.}
\Crefname{section}{Section}{Sections}
\Crefname{table}{Table}{Tables}
\crefname{table}{Tab.}{Tabs.}

\def\cvprPaperID{781} 
\def\confName{CVPR}
\def\confYear{2023}

\newcommand{\OURS}{InstMove\xspace}

\begin{document}

\title{InstMove:
Instance Motion for Object-centric Video Segmentation
}

\author{
Qihao Liu*$^1$ \quad Junfeng Wu*$^2$ \quad Yi Jiang$^3$ \quad Xiang Bai$^2$ \quad Alan Yuille$^1$ \quad Song Bai$^3$\\
{\normalsize $^1$Johns Hopkins University \quad $^2$Huazhong University of Science and Technology \quad $^3$ByteDance}
}

\maketitle
\def\thefootnote{*}\footnotetext{First two authors contributed equally. Work done during an internship at ByteDance. The code and models are available for research purposes at \href{https://github.com/wjf5203/VNext}{https://github.com/wjf5203/VNext}.}

\begin{abstract}
Despite significant efforts, cutting-edge video segmentation methods still remain sensitive to occlusion and rapid movement, due to their reliance on the appearance of objects in the form of object embeddings, which are vulnerable to these disturbances. 
A common solution is to use optical flow to provide motion information, but essentially it only considers pixel-level motion, which still relies on appearance similarity and hence is often inaccurate under occlusion and fast movement.
In this work, we study the instance-level motion and present \OURS, which stands for \textbf{Inst}ance \textbf{M}otion for \textbf{O}bject-centric \textbf{V}ideo S\textbf{e}gmentation. 
In comparison to pixel-wise motion, \OURS mainly relies on instance-level motion information that is free from image feature embeddings, and features physical interpretations, making it more accurate and robust toward occlusion and fast-moving objects.
To better fit in with the video segmentation tasks, \OURS uses instance masks to model the physical presence of an object and learns the dynamic model through a memory network to predict its position and shape in the next frame.
With only a few lines of code, \OURS can be integrated into current SOTA methods for three different video segmentation tasks and boost their performance.
Specifically, we improve the previous arts by 1.5 AP on OVIS dataset, which features heavy occlusions, and 4.9 AP on YouTubeVIS-Long dataset, which mainly contains fast moving objects.
These results suggest that instance-level motion is robust and accurate, and hence serving as a powerful solution in complex scenarios for object-centric video segmentation. 

\end{abstract}

\section{Introduction}
\label{sec:intro}

\begin{figure}[tb]
\centering
\includegraphics[width = 0.5\textwidth]{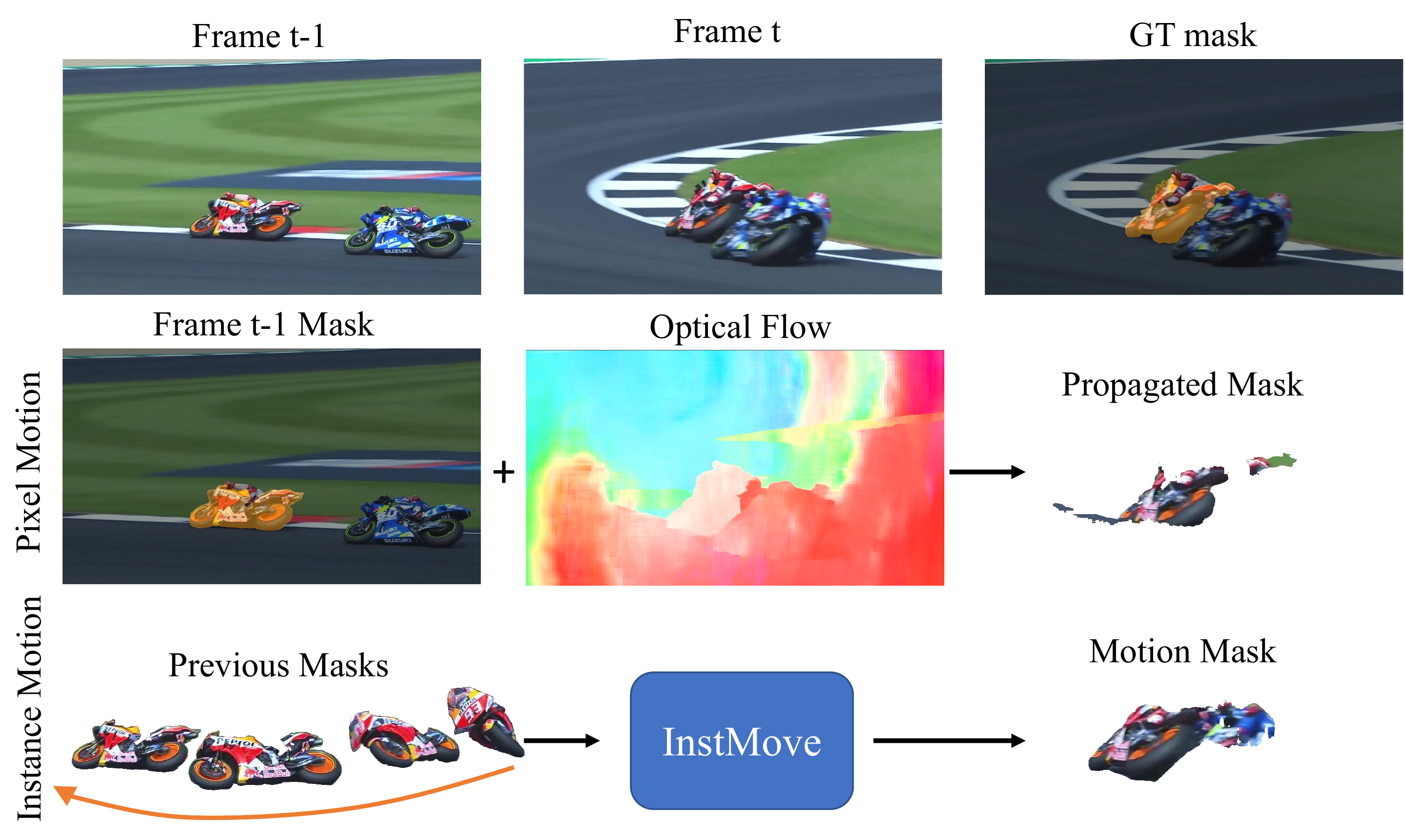}
\caption{
Different from optical flow that estimates pixel-level motion, \OURS learns instance-level motion and deformation directly from previous instance masks and predicts more accurate and robust position and shape estimates for the current frame, even in scenarios with occlusions and rapid motion.
}
\label{fig:intro}
\end{figure}


Segmenting and tracking object instances in a given video is a critical topic in computer vision, with various applications in video understanding, video editing, autonomous driving, augmented reality, etc. 
Three representative tasks include video object segmentation (VOS), video instance segmentation (VIS), and multi-object tracking and segmentation (MOTS). 
These tasks differ significantly from video semantic segmentation~\cite{shelhamer2016clockwork,hu2020temporally,li2018low,garcia2018survey}, which aims to classify every pixel in a video frame, hence we refer to them as object-centric video segmentation in this paper.
Despite significant progress, state-of-the-art (SOTA) methods are still struggle with occlusion, rapid motion, and significant changes in objects, resulting in a marked drop in handling longer or more complex videos.

One reason we observe is that most methods rely solely on appearance to localize objects and track them across frames. 
Specifically, a majority of VOS methods~\cite{STM,lu2020video,seong2020kernelized,li2020fast,xie2021efficient,STCN} use the previous frames as target templates and construct a feature memory bank of embeddings for all target objects. This is then used to match the pixel-level feature in the new frame. 
Online VIS~\cite{MaskTrackRCNN,sipmask,compfeat,STMask,CrossVIS,MinVIS,IDOL} and MOTS~\cite{pcan,unicorn} methods directly perform per-frame instance segmentation based on image features and use the object embeddings to track them through the video.
While these paradigms work well on simple videos, they are sensitive to intense appearance changes and struggle with handling multiple object instances with similar appearances, resulting in large errors when dealing with complex scenarios with complex motion patterns, occlusion, or deformation.

Apart from appearance cues, object motion, which is another crucial piece of information provided by videos, has also been extensively studied for video segmentation. The majority of motion models in related fields fall into two categories: One line of work uses optical flow to learn pixel-level motion. However, this approach does not help solve the problem of occlusion or fast motion since flow itself is often inaccurate in these scenarios~\cite{tsai2016video,oh2018fast}. The main reason causing the failure we argue is that optical flow still heavily relies on appearance cues to compute the pixel-level motion across the frame. The other line of work uses a linear speed model, which helps alleviate these tracking problems caused by occlusion and fast motion in MOT~\cite{SORT, DeepSORT, CTracker, bytetrack}. However, it oversimplifies the problem and thus provides limited benefits in other tasks such as VOS and VIS.

In this work, we aim at narrowing the gap between the two aforementioned lines of work by reformulating the motion module and providing \OURS, a simple yet efficient motion prediction plugin that enjoys the advantages of both solutions. First, it is portable and is compatible with and beneficial to approaches of video segmentation tasks. More importantly, similar to optical flow, it also provides high-dimensional information of position and shape, which can be beneficial for a range of downstream tasks in a variety of ways, and, similar to the dynamic motion model, it learns physical interpretation to model motion information, improving robustness toward occlusion and fast motion.

To achieve our objective, we utilize an instance mask to indicate the position and shape of a target object, and provide an RNN-based module with a memory network to extract motion features from previous masks, store and retrieve dynamic information, and predict the position and shape information of the next frame based on motion cues. However, while being robust towards appearance changes, predicting shape without the object appearance or image features results in an even less accurate boundary in simple cases. To solve this, we incorporate the low-level image features at the end of \OURS. Finally, to prove the effectiveness of \OURS on object-centric video segmentation tasks, we present two simple ways to integrate \OURS into recent SOTA methods in VIS, VOS, and MOTS, which improve their robustness with minimal modifications.

In the experiments section, we first validate that our motion module is more accurate and compatible with existing methods compared with learning motion and deformation with optical flow methods such as RAFT~\cite{raft}. 
We also show that it is more robust to occlusion and rapid movements.
Then we demonstrate the improvement of integrating our motion plugin into all recent SOTA methods in VOS, VIS, and MOTS tasks, particularly in complex scenarios with heavy occlusion and rapid motion. 
Remarkably, with only a few lines of code, we significantly boost the current art by 1.5 AP on OVIS~\cite{ovis}, 4.9 AP on YouTubeVIS-Long~\cite{ytvis22dataset}, and reduce IDSw on BDD100K~\cite{bdd100k} by 28.6\%.

In summary, we have revisited the motion models used in video segmentation tasks and propose \OURS, which contains both pixel-level information and instance-level dynamic information to predict shape and position. It provides additional information that is robust to occlusion and rapid motion. The improvements in SOTA methods of all three tasks demonstrate the effectiveness of incorporating instance-level motion in tackling complex scenarios.

\section{Related work}
\label{sec:related-work}

\subsection{Video Segmentation and Tracking}
Video segmentation and tracking is an important field including tasks such as MOT, MOTS, VIS, VOS, etc.
Multi-Object Tracking (MOT) aims to estimate the trajectories of multiple objects of interest in videos.
Dominant MOT methods~\cite{wang2020towards,qdtrack,zhou2020tracking,fairmot} mainly follow the tracking-by-detection paradigm, treating object detection and Re-ID as two separate tasks. 
MOTS is extended from MOT by changing the form of boxes to a fine-grained representation of masks.
MOTS benchmarks~\cite{bdd100k,mots} are typically drawn from the same scenarios as those of MOT~\cite{bdd100k,mot16}, and many MOTS methods~\cite{pcan,unicorn} are developed upon MOT trackers. 

Video Instance Segmentation (VIS)~\cite{MaskTrackRCNN,ovis} is an emerging vision task that aims to detect, classify, segment, and track object instances in videos simultaneously. It has similar settings to MOTS, but the videos of VIS are mainly from daily life, so there are a large variety of object categories and forms of motion, resulting in more complex scenarios compared with MOTS.
Current VIS methods can be categorized as offline or online methods. Offline methods~\cite{MaskProp,ProposeReduce,STEmSEG,VisTR,IFC,seqformer,VITA} take the whole video or a video clip as input and predict the mask sequence with a single step.
MaskProp~\cite{MaskProp} and Propose-Reduce~\cite{ProposeReduce} perform mask propagation in a video clip to improve mask and association. VisTR~\cite{VisTR} and related methods~\cite{IFC,seqformer,VITA} adopt the transformer~\cite{transformers} to VIS and model object queries for the input video.
Online methods~\cite{MaskTrackRCNN,sipmask,compfeat,STMask,CrossVIS,MinVIS,IDOL} perform instance segmentation on each frame of the video while tracking instances and optimizing results across frames by object embedding similarity and other cues. 

Video Object Segmentation (VOS) aims to segment video sequences, classifying each pixel corresponding to foreground objects in every frame.
We focus on a semi-supervised VOS setting, where the instance annotations of the first frame are provided.
Propagation-based methods~\cite{luiten2018premvos,chen2020state} refine the target segmentation mask in a temporal label propagation manner. 
Space-Time Memory (STM) network~\cite{STM,lu2020video,seong2020kernelized,li2020fast,xie2021efficient,STCN} is another popular framework that builds a memory bank for each object in the video and matches every query frame to the memory bank to readout memory for tracking and segmentation.

All these tasks focus on objects and instance tracking in the video, and hence object motion information is critical, particularly in the case of fast movement, occlusion, and large deformation. We aim to model instance-level motion to alleviate the problem caused by complex scenarios.

\subsection{Object Motion}
Object motion information in the video is a critical problem that is useful for a variety of tasks such as MOT, VOS, and VIS.
In MOT tasks, many methods try to utilize motion in various forms. Motion can be modeled for trajectory prediction by constant velocity assumption (CVA)~\cite{choi2010multiple,andriyenko2011multi}.
In crowded environments, pedestrian motion becomes more complex, prompting researchers to utilize the more expressive Social Force Mode~\cite{pellegrini2009you,scovanner2009learning,yamaguchi2011you ,alahi2016social,robicquet2016learning}.
Some methods~\cite{SORT, DeepSORT, CTracker, bytetrack} use the Kalman filter motion model to improve the robustness to serious occlusions and short-term disappearing followed by quick reappearance. 
Kalman filter is used to predict the location of the tracklets in the subsequent frame, and then the assignment cost matrix is computed as the distance between the predicted motion and all predicted bounding boxes. However, in MOT, motion is only modeled as the velocity and scale of the object. The fine-grained segmentation information of the object is hard to use when it comes to the extended task, \ie, MOTS.

Optical flow estimation has been extensively studied~\cite{flownet,flownetv2,raft} and it is widely used as motion information for VOS tasks.
Some methods~\cite{tsai2016video,hu2018unsupervised,xiao2018monet} estimate object segmentation and optical flow synergistically and reinforce the representation of the target frame by aligning and integrating representations from its neighbors such that the combination improves robustness to motion blur, appearance variation, and deformation of video objects. 
A commonly used architecture is dual-branch CNN models~\cite{cheng2017segflow,dutt2017fusionseg,tokmakov2017learning,li2019motion,ji2021full,yang2021learning}, consisting of an appearance branch and a motion branch.
They take RGB frames and optical flow as input and address the video object segmentation problem by leveraging the complementarity of appearance and motion cues.
Optical flow can be used to propagate an object segmentation over time but unfortunately, it is often inaccurate, particularly around object boundaries. 
To the best of our knowledge, no previous VIS algorithm explicitly models object motion and deformation. In this paper, we focus on the online VIS models and improve their performance in complex scenarios by integrating them with motion information.

Different from the previous motion methods, we propose instance-level motion that has both fine-grained information and physical meaning about an object compared with the velocity model and optical flow. 
We find that instance-level motion is more accurate and robust to occlusion and fast-moving objects, and can be integrated into current SOTA methods to boost their performance in complex scenarios.


\begin{figure*}[tb]
\centering
\includegraphics[width=0.95 \linewidth]{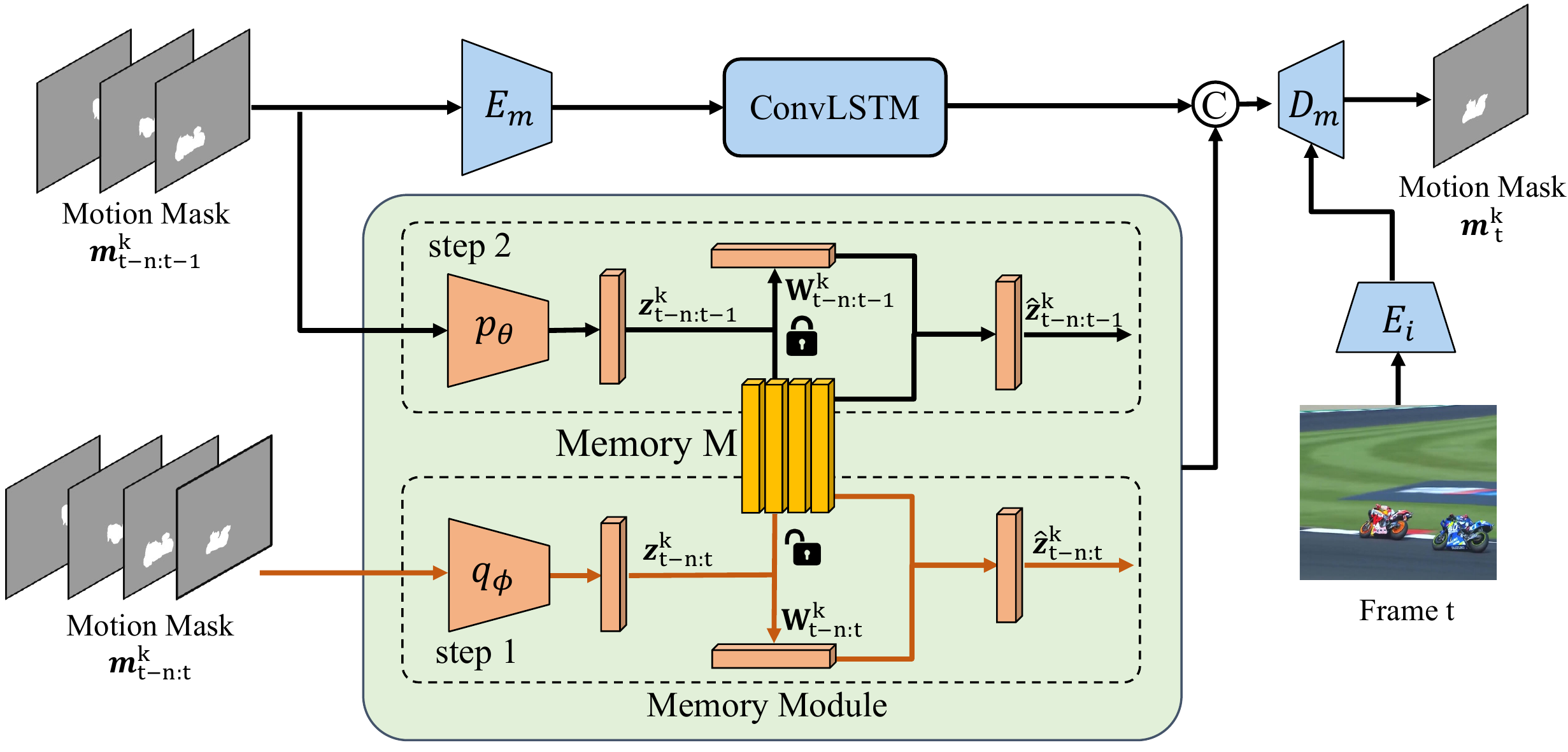}
\caption{\textbf{\OURS model pipeline.} We use instance mask $\textbf{m}^k$ to represent the position and shape of instance $k$. During the two-step training process, ground-truth masks $\textbf{m}_{t-n:t}^k$ and $\textbf{m}_{t-n:t-1}^k$ are provided and the memory bank $\textbf{M}$ is only updated in step 1. During inference, we directly use the estimated masks $\textbf{m}_{t-n:t-1}^k$ of the target video segmentation method as input to predict the mask $\textbf{m}_{t}^k$, and only step 2 is involved. Image features are added at the very end of \OURS to refine the boundary. }
\label{fig:pipeline}
\vspace{-2ex}
\end{figure*}

\section{\OURS: Instance-level Motion Prediction}
\label{sec:method}
We aim to design a flexible and efficient motion prediction module that can be easily added on top of any existing method. To this end, we design a memory-based motion prediction. We first formulate the problem in Sec.~\ref{sec:formulation}. Then we explain how we design the memory network (Sec.~\ref{sec:memory}) and how we use it to predict motion (Sec.~\ref{sec:motionPrediction}). Finally, we explain how we add the motion module to all methods in detail (Sec.~\ref{sec:plugin}).

\subsection{Problem Formulation and Motivation}
\label{sec:formulation}

Since we focus on segmentation tasks, we directly use the binary mask of an instance to represent the shape and position and predict motion upon it. 
Let $I_t\in \mathbb{R}^{w\times h\times 3}$ denotes the $t$-th frame of the input video and $\textbf{m}_t^k\in \mathbb{R}^{w\times h}$ denotes the binary mask of instance $k$ in this frame. The goal of this module is to learn the motion and deformation from $\textbf{m}_{t-n}^k$ to $\textbf{m}_{t-1}^k$, and then predict the shape and position, denoted by $\textbf{m}_{t}^k$. In our problem, we only require our module to predict one frame forward, but we wish it can take various number of frames as input, \ie $\textbf{m}_{t}^k = F(\textbf{m}_{t-n:t-1}^k), \forall n \geq 2$ where $F$ denotes our motion module.

A naive solution would be to use recurrent neural networks (RNNs) such as Conv-LSTM~\cite{shi2015convolutional} to extract the dynamics from the input frames and predict the next, but this solution yields poor results. They tend to make pixel-wise predictions for the entire image, while we want to learn instance-level motion and thus all pixels should be considered as a unit.
Following the concept of conditional variational autoencode (CVAE) that models the dynamics with a Gaussian and samples latent variables $\textbf{z}_t$ from this distribution for prediction~\cite{ling2020character,rempe2021humor}, we also use an encoder during training to learn the motion pattern represented by $\textbf{z}_t$. However, we maintain a memory network $\textbf{M}$ to store representative motion patterns. The stored motion patterns are then used during inference to help refine the incomplete pattern that is extracted only from previous frames $\textbf{m}_{t-n:t-1}^k$, and the refined pattern, which contains the instance-level motion information, is used to assist the prediction of the next frames $\textbf{m}_{t}^k$. Fig.~\ref{fig:pipeline} shows the overall framework of our motion prediction plugin.

\subsection{Memory Network}
\label{sec:memory}

During training, we assume that we have access to the target mask $\textbf{m}_{t}^k$. We can learn a motion feature/pattern $\textbf{z}_{t-n:t}^k = q_{\phi}(\textbf{m}_{t-n:t}^k)$. Different from the CVAE model that uses $q_{\phi}$ to approximate a posterior distribution and use a network to learn the mean and variance, we directly use $q_{\phi}$ to regress a vector of length $l$ here. The motion pattern $\textbf{z}_{t-n:t}^k\in \mathbb{R}^{l}$ is then stored in a memory bank $\textbf{M}\in \mathbb{R}^{c\times l}$ by a corresponding attention weight vector $\textbf{w}_{1:t}^k\in\mathbb{R}^{c}$ following a typical solution~\cite{gong2019memorizing,pei2019memory,lee2021video}. Specifically, let $\textbf{v}_i\in \mathbb{R}^{l},i=1,...,c$ denotes memory vectors of the memory bank $\textbf{M}$, then the $i$-th element of $\textbf{w}_{t-n:t}^k$ is computed via a soft-max function:
\begin{align}
    w_{1:t,i}^k = \frac{exp\{S_{cos}(\textbf{z}_{t-n:t}^k,\textbf{v}_i)\}}{\sum_{j=1}^cexp\{S_{cos}(\textbf{z}_{t-n:t}^k,\textbf{v}_j)\}}
    \label{eq:1}
\end{align}
where $S_{cos}(\textbf{a},\textbf{b})=\frac{\textbf{a}\cdot \textbf{b}}{||\textbf{a}||\cdot ||\textbf{b}||}$ is the cosine similarity between vector $a$ and $b$. The weight vector $\textbf{w}_t^k$ is a non-negative weight vector with entries sum to one, and it enables us to access the memory bank. Given a latent variable $\textbf{z}\in \mathbb{R}^{l}$, we can obtain the corresponding memory feature:
\begin{align}
    \hat{\textbf{z}} = \textbf{w}\textbf{M}=\sum_{i=1}^cw_i\textbf{v}_i
     \label{eq:2}
\end{align}
This parameters of $\textbf{M}$ are updated via backpropagation. This process provides us a explicit way to store different motion patterns and to retrieve a given motion pattern $\textbf{z}_{t-n:t}^k$ from the memory network.

Given a trained memory bank $\textbf{M}$ and the input masks $\textbf{m}_{t-n:t-1}^k$, we can use another encoder $p_\theta$ to extract motion patterns $\textbf{z}_{t-n:t-1}^k = p_\theta(\textbf{m}_{t-n:t-1}^k)$. Note that $p_\theta$ and $q_\phi$ share the same architecture but use different parameters. Then we can use Eq.~\ref{eq:1} and ~\ref{eq:2} to access the memory bank, match $\textbf{z}_{t-n:t-1}^k$ with a learned motion pattern, and retrieve the corresponding motion pattern/prior $\hat{\textbf{z}}_{t-n:t-1}^k$. The memory bank $\textbf{M}$ is not updated in this step.

\subsection{Motion Prediction with Memory Network}
\label{sec:motionPrediction}
With the help of memory network, we can then use an RNN-based network to predict the target frame $\textbf{m}_{t}^k$. Specifically, given the input masks $\textbf{m}_{t-n:t-1}^k$, we use a mask encoder $E_m$ followed by a Conv-LSTM $L$ to extract mask features $\textbf{f}_{t-n:t-1}^k=L(E_m(\textbf{m}_{t-n:t-1}^k))$. Then the mask feature $\textbf{f}_{t-n:t-1}^k$ and the retrieved motion pattern $\hat{\textbf{z}}_{(\cdot)}^k$ are combined together~\cite{lee2021video} and fed to a mask decoder $D_m$ to predict the target mask $\textbf{m}_{t}^k=D_m(\textbf{f}_{t-n:t-1}^k,\hat{\textbf{z}}_{(\cdot)}^k)$. 

For each iteration, the training process involves two steps. In step 1, we train the encoder $q_{\phi}$ and the parameters of memory bank $\textbf{M}$ with input $\textbf{m}_{t-n:t}^k$. Then $\hat{\textbf{z}}_{(\cdot)}^k=\hat{\textbf{z}}_{(t-n:t)}^k$ is used to predict the target mask $\textbf{m}_{t}^k$. In step 2, we freeze the the parameters of $\textbf{M}$ and feed $\textbf{m}_{t-n:t-1}^k$ to encoder $p_{\theta}$. We only train the encoder $p_{\theta}$ in this step and $\hat{\textbf{z}}_{(\cdot)}^k=\hat{\textbf{z}}_{(t-n:t-1)}^k$ is used for prediction. At test time, only $p_{\theta}$ with input $\textbf{m}_{t-n:t-1}^k$ is used to compute the latent variable and retrieve motion pattern from the memory network.

\noindent{\textbf{Refine Mask Boundary with Image Features.}} 
To obtain more accurate shape estimates, we consider using the image feature at the very end to refine $\textbf{m}_{t}^k$. We employ the first two stages of ResNet-50 as image encoder $E_i$ to extract low-level features from the original image, which generate feature maps with 8 strides and 4 strides, respectively.
They are then upsampled by two and added together with the motion feature in the decoder. 
Note that this step is optional, and directly using the outputs of the backbone of original video segmentation methods could also yield comparable or even larger improvements (See Sec.~\ref{subsec:Ablation}).

\noindent{\textbf{Loss Function.}}
As the predicted motion $\textbf{m}_{t}^k$ is in the form of binary masks, we use the mask loss $\mathcal{L}_{\text{mask}}$ that commonly used in segmentation methods such as DETR~\cite{carion2020end} to supervise the learning of motion. It is defined as a combination of the Dice~\cite{milletari2016v} and Focal loss~\cite{lin2017focal}: 
$ \mathcal{L}_{\text{mask}} = \lambda_{focal}\mathcal{L}_{focal} + \lambda_{dice}\mathcal{L}_{dice}$.
We set $\lambda_{focal}=1$ and $\lambda_{dice}=5$.

\subsection{Plugin for Object-centric Video Segmentation}
\label{sec:plugin}
We consider three different object-centric video segmentation tasks and provide two straightforward ways to utilize our motion module: (1) To help the tracking process, we use motion prediction to obtain a motion score and combine it with the original matching score of embedding similarities. (2) To improve the segmentation quality, we use the motion mask as an attention map and concatenate it with the feature map in the decoder. Both can be easily implemented in a few lines of code, but improving the performance of all methods in these three tasks, especially in complex scenarios with occlusion or fast-moving objects, demonstrating the efficacy of using the instance-level motion.

\noindent{\textbf{VIS.}}
For the VIS task, we adopt CrossVIS~\cite{CrossVIS}, a classic method, as well as two recently proposed SOTA methods, MinVIS~\cite{MinVIS} and IDOL~\cite{IDOL}, as our baselines.  
To eliminate the influence of random variation during training, we modify only the inference stage of the VIS models and utilize the official pre-trained weights for inference.

1) CrossVIS compares the cosine similarity of embeddings and combines the cues of the box position and the classification results to obtain a matching score, which is used to assign the object in the current frame to the existing tracklets.
Therefore, we use the previous masks in the existing tracklets to predict the motion masks of the objects in the current frame, then we calculate the mask IoUs between motion masks and the instance segmentation results as motion score. Finally, we add the motion score to the original matching score to use motion to help with the tracking.

2) IDOL only uses contrastive embedding to calculate the bi-softmax score to assign the objects in the current frame, so we add the motion score to it in the same way.

3) MinVIS employs the Hungarian algorithm on a score matrix S, the cosine similarity of query embeddings, for tracking. To avoid introducing redundant information, we only calculate the motion score for the top 20 tracklets with the highest confidence scores.

\noindent{\textbf{MOTS.}}
MOTS has a similar setting to VIS, we adopt PCAN~\cite{pcan} and Unicorn~\cite{unicorn}. They use the bi-softmax between object embeddings as the matching score to achieve object tracking. To improve tracking quality, we add motion scores in the same way to introduce motion information.

\noindent{\textbf{VOS.}}
For the VOS task, we adopt STCN~\cite{STCN} as an example. Different from previous algorithms for VIS and MOTS, STCN constructs a memory bank for each object in the video, and when predicting the next frame, it needs to read the features in memory to decode them into the predicted mask.
Therefore, we added a convolutional layer into the decoder to process the concatenation of the motion mask and feature map. We retrain the STCN to adapt the input of motion information and load the frozen motion module to generate a motion mask during training.

\section{Experiments}
\label{sec:experiments}

\begin{table*}[!t]
\centering
\small
\begin{tabular}{lcccccccccccc}
\toprule
\multirow{2}*{Method} &\multicolumn{6}{c}{OVIS} &\multicolumn{6}{c}{OVIS-Sub-Sparse}  \\
 \cmidrule(lr){2-7} \cmidrule(lr){8-13} 
  &$\rm AP$  &$\Delta_{AP}$   &$\rm AP_{50}$   &$\rm AP_{75}$ &$\rm AR_{1}$  &$\rm AR_{10}$  &$\rm AP$ &$\Delta_{AP}$    &$\rm AP_{50}$  &$\rm AP_{75}$ &$\rm AR_{1}$  &$\rm AR_{10}$  \\
\midrule
 CrossVIS~\cite{CrossVIS} &12.6 &  &28.4 &10.8 &8.9 &17.1  &7.0  &  &16.8  &5.5  &6.1  &11.0   \\
 CrossVIS+\OURS &16.7 &\textcolor{red}{+ 4.1}  &35.1  &15.0  &10.0  &21.6  &8.9  &\textcolor{red}{+ 1.9}  &20.7  &7.0  &7.4  &13.3   \\
\midrule
 MinVIS~\cite{MinVIS} &26.2 &  &48.2  &25.0  &14.4  &30.1  &15.3  &  &31.8  &13.6  &10.1  &20.6   \\
 MinVIS+\OURS &27.6 &\textcolor{red}{+ 1.4}  &51.0  &26.4  &14.4  &31.5  &18.2  &\textcolor{red}{+ 2.9}  &36.9 &16.0  &11.2  &23.3   \\
\midrule
 IDOL~\cite{IDOL} &29.2  &  &49.8  &29.1  &14.8  &37.0  &16.5  & &34.0  &15.3  &10.3  &25.9   \\
 IDOL+\OURS &30.7  &\textcolor{red}{+ 1.5}   &51.4  &30.9  &15.0  &37.7  &18.5  &\textcolor{red}{+ 2.0}  &37.8  &16.8  &10.7  &27.0   \\
\bottomrule
\end{tabular}
\caption{\textbf{Quantitative results of video instance segmentation on OVIS and OVIS-Sub-Sparse validation set.} 
We directly test the official pre-trained models for MinVIS and IDOL, and only use our motion module during inference, with no additional training. 
For CrossVIS, the pre-trained model is not available, so we report the performance of the model trained by ourselves with the official code.
The motion modules used in this table are trained on OVIS and OVIS-Sub-Sparse datasets, respectively.}
\label{OVIS_Sparse}
\vspace{-2ex}
\end{table*}

\subsection{Datasets and Metrics}
\label{subsec:datasets}
\noindent{\textbf{VIS.}}
For VIS task, the OVIS~\cite{ovis}, YouTube-VIS-Long~\cite{ytvis22dataset}, OVIS-Sub, and OVIS-Sub-Sparse datasets are used.

OVIS is a recent and challenging dataset, with 607 training videos, 140 validation videos, and 154 test videos. The videos are considerably longer and last 12.77s on average. OVIS is characterized by challenging scenarios, featuring severe occlusions, complex motion patterns, and rapid object deformation. 
YouTube-VIS-Long tackles the scenarios of long and highly complicated sequences, consisting of 71 long videos in the validation set. Compared with the previous YouTube-VIS dataset, the videos are longer and with only 1 FPS sample rate, making it very challenging.
OVIS-Sub is a subset of the OVIS dataset. 
The annotations for the OVIS validation set is not available, so we split the training set into custom training and validation sets to evaluate the predicted motion mask. We divided the 607 training videos into a validation subset of 122 videos and a new training subset, following the approach of IDOL~\cite{IDOL}.

Handling longer videos with lower sampling rates is a valuable scenario, but the small scale of YouTube-VIS-Long makes the analysis on it less convincing. To address this, we sparsely sample the videos from OVIS-Sub to make a larger dataset with the same features, named OVIS-Sub-Sparse. 
For the validation subset, the frames and annotations are kept every 5 frames to achieve a sampling rate of about 1FPS, similar to YouTube-VIS-Long.

\noindent\textbf{MOTS.}
We evaluate the methods on BDD100K~\cite{bdd100k}, which is a challenging self-driving dataset with 154 videos for training, 32 videos for validation, and 37 videos for testing. The dataset provides 8 annotated categories for evaluation, where the images in the tracking set are annotated per 5 FPS with 30 FPS frame rate. 
We report multi-object tracking and segmentation accuracy (MOTSA), number of Identity Switches (IDSw), and ID F1 score. ID F1 score is the ratio of correctly identified detection over the average number of ground-truths and detection.

\noindent\textbf{VOS.}
DAVIS-17~\cite{davis17} contains 30 videos in the validation set and there could be multiple tracked targets in each sequence. 
YouTube-VOS 2019~\cite{ytvos2018} is a large-scale benchmark for multi-object video segmentation, providing 3,471 videos for the training (65 categories) and 507 videos (65 training categories, 26 unseen categories ) for the validation.
For DAVIS 2017, we report standard metrics~\cite{davis16}: Jaccard index  $\mathcal{J}$, contour accuracy $\mathcal{F}$ and their average $\mathcal{J\&F}$.
For YouTubeVOS, we report $\mathcal{J}$ and $\mathcal{F}$ for both seen and unseen categories, and the averaged overall score $\mathcal{G}$.

\subsection{Implementation Details}

\noindent{\textbf{Model Setting.}}
All encoders and decoders involved consist of several Conv layers or fully connected layers. Three 3D-Conv layers are used for $q_{\phi}$ and $p_{\theta}$, and two 2D-Conv layers are used for mask encoder $E_m$, followed by a 3-layer ConvLSTMs. For mask decoder $D_m$, we employ three 2D-DeConv layers. The image encoder $E_i$ uses the first two layers of a pre-trained ResNet-50 backbone.

\noindent{\textbf{Training and Inference.}}
For evaluation on these video segmentation tasks, we first train \OURS on the corresponding dataset independently and freeze the \OURS weights during inference.
We re-scale all the input image masks to 384×384 with padding to preserve the aspect ratios. During training, the ground-truth masks are used, and during inference, we use the masks predicted by the target method as the input of \OURS. To be able to handle different length of inputs, we set randomly select $n$ from $[2,5]$ during training.

\subsection{Comparison to Optical Flow}
\label{subsec:raft}

In this section, we demonstrate that our motion module is more accurate and suitable for object-centric video segmentation tasks than the commonly used paradigm of learning pixel-wise motion with optical flow methods such as RAFT. We consider two different scenarios: videos with occlusions and videos with fast-moving objects, and conduct experiments on the OVIS-Sub, and OVIS-Sub-Sparse datasets.

For fair comparisons, we first use the official pre-trained model of RAFT~\cite{raft} to predict optical flow on the training split of each dataset. The predictions are then used as pseudo-labels to fine-tune the RAFT model for 50k iterations (following the original RAFT paper). After that, we use the new model to generate optical flow predictions on val split and propagate the masks based on them. For our motion module, we directly train it from scratch on the training split for 100K iterations. The results are shown in Table~\ref{table:Raft}.
We can see that our model outperforms the RAFT-based motion module by a significant margin, especially for complex videos with occlusion or fast-moving objects.

\begin{table}
\renewcommand\arraystretch{0.9}
    \small 
    \centering
    \resizebox{0.9\columnwidth}{!}{
    \begin{tabular}{lcccc}
    \toprule
    &\multicolumn{2}{c}{OVIS-Sub} & \multicolumn{2}{c}{OVIS-Sub-Sparse}\\
    \cmidrule(lr){2-3} \cmidrule(lr){4-5}
    & mAP($\uparrow$) & IOU($\uparrow$) & mAP($\uparrow$) & IOU($\uparrow$)\\
    \midrule
    RAFT~\cite{raft} & 46.3 & 64.2 & 30.6 & 49.5 \\
    \OURS (Ours) & \textbf{67.8} & \textbf{79.2} & \textbf{57.3} & \textbf{70.7} \\
    \bottomrule
    \end{tabular}}
    \caption{\textbf{Comparison for motion prediction using our method and optical flow.} We compare the propagated masks from RAFT and the predicted masks from our motion module here. All models are trained/fine-tuned on the corresponding training split and tested on the test split with an image size of $384\times 384$. Our model is more accurate and robust under occlusion and fast motion.}
    \label{table:Raft}
    \vspace{-3mm}
\end{table}

\begin{figure*}[tb]
\centering
\includegraphics[width=0.9 \linewidth]{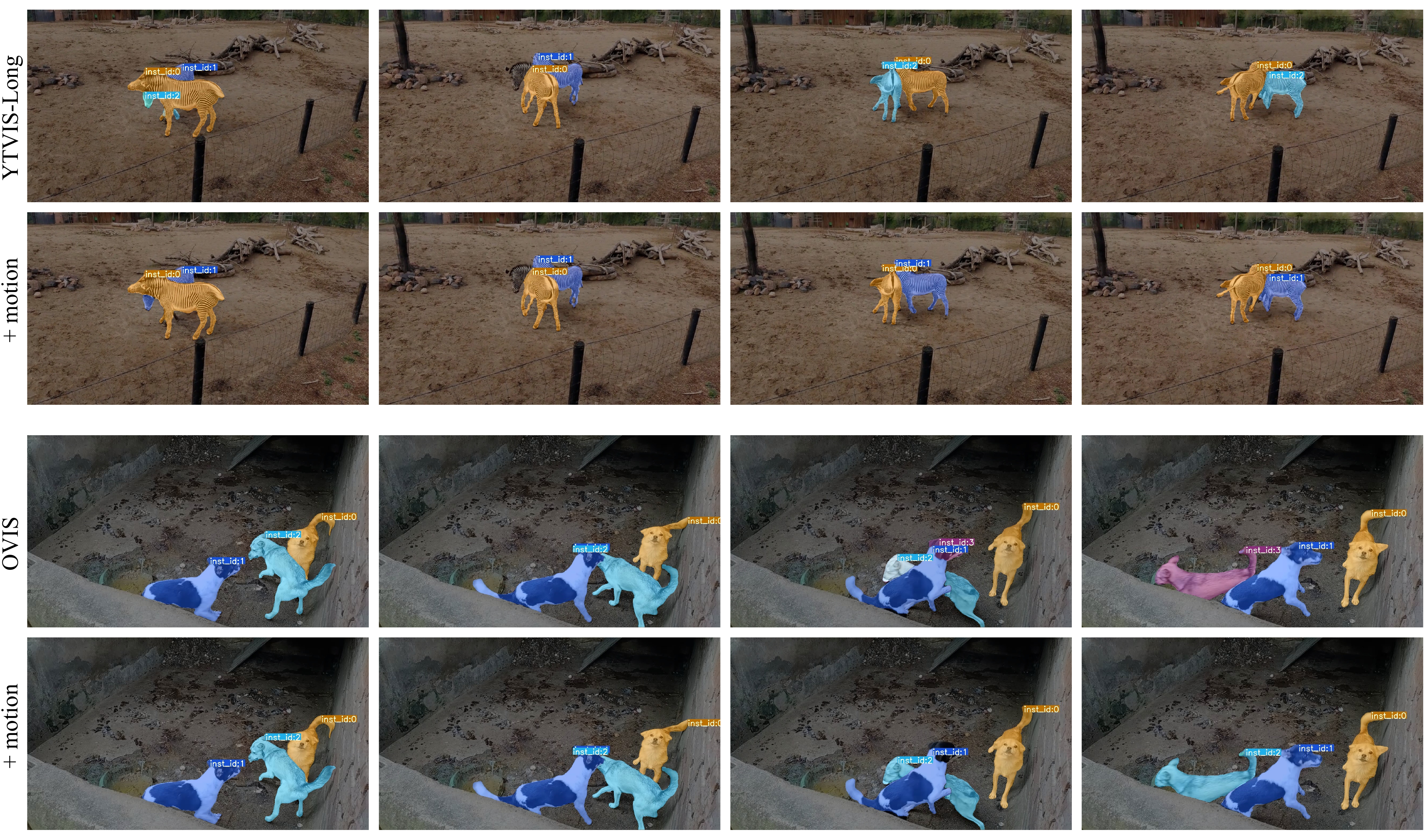}
\caption{
\textbf{Qualitative results of IDOL on the YouTube-VIS 2022 Long and OVIS validation dataset.} We can see that the ID switch happens in the third and last frame of the first row, probably because these two zebras are very close and have similar appearances.
The same error happens in the last frame of the third row.
Leveraging motion information effectively solves this problem and makes IDOL more robust toward occlusion and similar objects.
}
\vspace{-1ex}
\label{fig:visualization}
\vspace{-1ex}
\end{figure*}

\subsection{Evaluations on Video Instance Segmentation}
\label{subsec:VIS}
In this section, we demonstrate that our motion module is able to boost most VIS methods on three challenging datasets.
We plug the motion module into three online VIS algorithms to improve their tracking results as described in Sec.~\ref{sec:plugin}, and then evaluate on three challenging VIS datasets.
For IDOL and MinVIS, we directly download the pre-trained model weight on the OVIS and YTVIS21 datasets from the official model zoo. 
For the OVIS-Sub-Sparse dataset, we follow the official training scripts to train new IDOL and MIinVIS models on the training subset.
Since there are no available pre-trained models of CrossVIS in the official model zoo, we retrained CrossVIS on these three datasets.
All algorithms use ResNet-50 as backbone.

\begin{table}
\renewcommand\arraystretch{0.9}
    \small 
    \centering
    \resizebox{\columnwidth}{!}{
    \begin{tabular}{*6{l}}
    \toprule
    &$\rm AP$   &$\rm AP_{50}$   &$\rm AP_{75}$ &$\rm AR_{1}$  &$\rm AR_{10}$  \\
    \midrule
    MinVIS~\cite{MinVIS} & 22.3  &44.6 &20.2 &19.2 &26.4 \\
    MinVIS+\OURS &\textbf{25.9}  &\textbf{46.7} &\textbf{24.8} &\textbf{24.5} &\textbf{30.0} \\
    \midrule
    IDOL~\cite{IDOL} &35.7  &62.4 &35.7 &32.0 &44.7\\
    IDOL+\OURS & \textbf{40.6} &\textbf{67.2} &\textbf{45.1} &\textbf{35.0} &\textbf{48.2} \\
    \bottomrule
    \end{tabular}}
    \caption{\textbf{Quantitative results of video instance segmentation on YouTube-VIS Long validation set.} The motion module used in this table is trained on YouTube-VIS 2021 training set.}
    \vspace{-1ex}
    \label{table:YTVISLong}
    \vspace{-3ex}
\end{table}

As shown in Table~\ref{OVIS_Sparse}, our motion module can improve the performance of CrossVIS by 4.1 AP, MinVIS by 1.4 AP, and IDOL by 1.5 AP on OVIS validation set. IDOL is the SOTA method for VIS tasks and has made a great improvement in tracking process. It uses a very sophisticated mechanism to solve the tracking problem. The improvement on IDOL shows that we can further improve the tracking quality of VIS by integrating the motion information.
When it comes to the YTVIS-Long dataset with lower FPS, it can be seen that the improvement brought by the motion module becomes more significant in Table~\ref{table:YTVISLong}. IDOL and MinVIS are improved by 4.9 AP and 3.6 AP respectively.
As shown in the Table~\ref{OVIS_Sparse}, MinVIS improves 2.9 AP on OVIS-Sub-Sparse, and IDOL improves 2.0 AP. The results on OVIS-Sub-Sparse demonstrate the improvement brought by our motion module on low FPS videos more convincingly.
The low FPS makes the fast-moving objects occur frequently, which is a very challenging scenario. 
In Figure~\ref{fig:visualization}, we provide qualitative results to demonstrate how \OURS improves the tracking result in occlusion and fast-moving scenarios.
We believe that the motion information can help the VIS algorithm to better utilize the motion information of the object and to make accurate tracking.

\begin{table}
\renewcommand\arraystretch{0.9}
    \small 
    \centering
    \resizebox{\columnwidth}{!}{
    \begin{tabular}{lccc}
    \toprule
    & mMOTSA ($\uparrow$) & IDF1 ($\uparrow$) & ID switch ($\downarrow$)\\
    \midrule
    PCAN~\cite{pcan} &30.5 &44.5  &775\\
    PCAN+\OURS & \textbf{30.8} & \textbf{44.7}  & \textbf{699} \\
    \midrule
    Unicorn~\cite{unicorn} & 30.7  &47.1   &2044 \\
    Unicorn+\OURS & \textbf{31.2}  &\textbf{48.2}  &\textbf{1460  }\\
    \bottomrule
    \end{tabular}}
    \caption{\textbf{Quantitative results of MOTS on BDD100K validation set.} \OURS is trained on BDD100K MOTS training set.}
    \label{table:MOTS}
 \vspace{-2ex}
\end{table}

\begin{table}[t]
\renewcommand\arraystretch{1.0}
    \small 
    \centering
    \resizebox{1.0\columnwidth}{!}{
    \begin{tabular}{lcccccc}
    \toprule
    \multirow{2}*{Unicorn} &\multicolumn{2}{c}{official val} & \multicolumn{2}{c}{occlusion-subset} & \multicolumn{2}{c}{continuous-subset} \\
    \cmidrule(lr){2-3} \cmidrule(lr){4-5} \cmidrule(lr){6-7}
    & IDSw($\downarrow$) & IDSw/I($\downarrow$) & IDSw & IDSw/I & IDSw & IDSw/I\\
    \midrule
    w/o InstMove & 2044 &41.3\%  & 1194 &129.6\%  &856 &21.3\%\\
    w/ InstMove & 1460 &29.5\%   & 882 &95.7\%  & 579  & 14.4\% \\
    \bottomrule
    \end{tabular}}
    \caption{\textbf{Comparison of IDSw on MOTS occlusion subsets.} We take BDD100K MOTS validation set as `official val', and split it into occlusion and continuous subsets.}
    \label{table:MOTS_subset}
    \vspace{-2ex}
\end{table}

\subsection{Evaluations on Multi-Object Tracking and Segmentation}
\label{subsec:MOTS}

To evaluate the improvement of our motion module on MOTS tasks, we choose the SOTA algorithms PCAN~\cite{pcan} and Unicorn~\cite{unicorn} as our baseline algorithms. As with the VIS algorithm, we directly use the official pre-trained models. \OURS only changes the matching score for tracking during inference, thereby improving the tracking performance of the models without changing their segmentation quality.
As shown in Table~\ref{table:MOTS}, integrating with InstMove improves tracking accuracy and reduces the IDSw of PCAN by 9.8\% and Unicorn by 28.6\%, thus improving the mMOTSA and IDF1 of them.

For objects that disappear and reappear in BDD100K, we filter them out and form the occlusion-subset, which contains 921 objects. The rest objects form the continuous-subset.
We evaluate the performance of Unicorn  w/ and w/o InstMove,
and report number of Identity Switches (IDSw) and IDSw per instance (IDSw/I).
As shown in Tab.~\ref{table:MOTS_subset}, the IDSw/I is reduced more significantly on the occlusion-subset, which indicates that InstMove is able to help MOTS algorithm solve the tracking problem caused by occlusion.

\begin{table}
\renewcommand\arraystretch{0.9}
    \small 
    \centering
    \resizebox{\columnwidth}{!}{
    \begin{tabular}{*4{c}}
    \toprule
 DAVIS 2017   & J\&F-Mean($\uparrow$) & J-Mean($\uparrow$) & F-Mean($\uparrow$)\\
    \midrule
    STCN~\cite{STCN} & 83.8 & 80.4 & 87.2 \\
    STCN+\OURS & \textbf{85.1} & \textbf{82.3} & \textbf{87.9} \\
    \bottomrule
    \end{tabular}}
    
    \vspace{1mm}
    
   \resizebox{\columnwidth}{!}{
    \begin{tabular}{*4{ccccc}}
    \toprule
  YouTubeVOS 2019   & $\mathcal{G}$ ($\uparrow$) &$\mathcal{J}_{s}$ ($\uparrow$) &$\mathcal{F}_{s}$ ($\uparrow$) &$\mathcal{J}_{u}$ ($\uparrow$) & $\mathcal{F}_{u}$ ($\uparrow$)\\
    \midrule
    STCN~\cite{STCN} & 82.8 & 81.5 & 77.9 & 85.8  &85.9 \\
    STCN+\OURS & \textbf{83.4 }&\textbf{ 82.5} & \textbf{77.9} & \textbf{86.9} & \textbf{86.0} \\
    \bottomrule
    \end{tabular}}
    
    \caption{\textbf{Quantitative results of video object segmentation on DAVIS 2017 and YouTube-VIS 2019 validation set.} 
    The motion module is first pre-trained on DAVIS and YouTube VOS dataset. Then we freeze the motion module and re-train the STCN with the same training setup as the baseline.}
    \label{table:VOS}
    \vspace{-2ex}
\end{table}

\begin{table}
\renewcommand\arraystretch{0.9}
    \small 
    \centering
    \resizebox{1\columnwidth}{!}{
    \begin{tabular}{lcccc}
    \toprule
    &\multicolumn{2}{c}{OVIS-Sub} & \multicolumn{2}{c}{OVIS-Sub-Sparse}\\
    \cmidrule(lr){2-3} \cmidrule(lr){4-5}
    & mAP($\uparrow$) & IOU($\uparrow$) & mAP($\uparrow$) & IOU($\uparrow$)\\
    \midrule
    No image encoder & 61.0 & 74.4 & 46.6 & 61.7 \\
    Fixed image encoder & 66.2 & 77.3 & 53.3 & 67.1 \\
    Full model & 67.8 & 79.2 & 57.3 & 70.7 \\
    \bottomrule
    \end{tabular}}
    \caption{\textbf{Effects of using image features during motion prediction.} For `fixed image encoder', we directly use the pre-trained weights of the first two layers of the image backbone in IDOL~\cite{IDOL}.}
    \label{table:ablImg}
    \vspace{-2ex}
\end{table}

\subsection{Evaluations on Video Object Segmentation}
\label{subsec:VOS}

Since we plug a convolution layer into the original STCN~\cite{STCN} model to introduce motion information from \OURS, we need to retrain the STCN model. We first reload the pre-trained weight that is obtained by training on static image datasets~\cite{wang2017learning,shi2015hierarchical,zeng2019towards,cheng2020cascadepsp,li2020fss} with synthetic deformation, following STCN. Then we perform main training of the same setup on YouTubeVOS~\cite{ytvos2018} and DAVIS~\cite{davis16,davis17}, and evaluate the 
performance on YouTube-VOS 2019 and DAVIS 2017 validation set.
As shown in Table~\ref{table:VOS}, integrated with our motion module, STCN can be improved from 83.8 to 85.1 on DAVIS2017, and from 82.8 to 83.4 on YouTubeVOS 2019.
This demonstrates that the estimation of the position and shape of the object in the next frame predicted by \OURS can be used as prior information to improve the segmentation ability of the VOS models.

\subsection{Ablation Study}
\label{subsec:Ablation}

\noindent\textbf{Refine Mask boundary with Image Features.} Tab.~\ref{table:ablImg} shows the effects of using image feature to refine the motion prediction. Directly adding the low-level image features can further improve the accuracy of motion prediction. Using a fixed image encoder also boosts performance. Since all methods require an image backbone to extract features, using a fixed image encoder does not incur extra costs. However, the motion module will no longer be universal, and we need to train a separate motion module for each model.

\noindent\textbf{Effect of Using Different Motion Modules.}
To investigate the impact of different motion modules on the performance of downstream tasks like VIS,
we use the motion module with a fixed image encoder to predict the motion score for IDOL.
As shown in Tab.~\ref{table:ablatioin_ovis},
freezing the image encoder still improves the tracking results.
To understand whether the Kalman filter motion model used by previous MOT methods~\cite{SORT, CTracker}, we employ a Kalman filter to predict the trajectory of the center of the bounding box, and then calculate the Euclidean distance between the trajectories and the predicted boxes to get the motion score for tracking. The results show that the Kalman filter motion model barely enhances the performance of VIS task.

\begin{table}
\renewcommand\arraystretch{0.9}
    \small 
    \centering
    \resizebox{1\columnwidth}{!}{
    \begin{tabular}{*6{c}}
    \toprule
    OVIS &$\rm AP$   &$\rm AP_{50}$   &$\rm AP_{75}$ &$\rm AR_{1}$  &$\rm AR_{10}$  \\
    \midrule
    Full model &30.7 &51.4  &30.9  &15.0  &37.7 \\
    Fixed image encoder & 30.9 &52.4 &30.8 &15.1 &38.0 \\
    No image encoder & 30.1 &51.1 &29.8 &14.7 &37.5 \\
    Kalman  &29.2 &49.7 &28.8 &15.1 &36.1  \\
    w/o motion &29.2 &49.8  &29.1  &14.8  &37.0   \\
    
    \bottomrule
    \end{tabular}}
    \caption{
    \textbf{Effects of fixed-encoder motion module and Kalman filter motion module in VIS task.}
    We report the results on OVIS validation set. For `fixed image encoder', the motion module shares the image encoder weights with IDOL.
    For `Kalman', we replace our motion module with a Kalman filter to predict the trajectory of the center of the bounding box.
    }
    \label{table:ablatioin_ovis}
    \vspace{-1ex}
\end{table}

\section{Conclusion}
In this work, we introduced \OURS that learns the dynamic model of an object by modeling its physical presence with instance masks. 
In contrast to the velocity model and optical flow-based motion model, \OURS has both fine-grained information and the physical meaning of an object. 
It provides additional information that is robust in complex scenarios and benefits the video segmentation methods.

\textbf{Acknowledgements.} QL acknowledges support from the Office of Naval Research with award N00014-21-1-2812. JW acknowledges support from the National Natural Science Foundation of China No. 62225603. We thank the reviewers for their efforts and valuable feedback to improve our work.

{\small
\bibliographystyle{ieee_fullname}
\bibliography{egbib}
}

\clearpage

\appendix

\section*{Appendices}
Here we provide implementation details (Sec.~\ref{supsec:imp}) and extended experimental results (Sec.~\ref{supsec:exp}) omitted from the main paper for brevity.
\section{Implementation Details}
\label{supsec:imp}
\noindent\textbf{Training \OURS.} We use Adam optimizer with a learning rate of $5\times 10^{-5}$ during training. For all experiments, the model is training for 10K iterations on 8 V100 GPUs of 32G RAM, with a batch size of 32. We re-scale all the input image masks to $384\times 384$ with padding to preserve the aspect ratios. We set memory length $l=256$ and memory size $c=100$. During training, we randomly select adjacent 3 to 5 frames (the last frame serves as the target frame) to enable the model to handle different input lengths during inference.

\section{Comparison with Optical Flow}
\label{supsec:exp}

Optical flow is used to provide motion information in many previous methods. Since it considers pixel-level motion, it can be used to propagate previous object masks to the current frame through a warp layer.
In this section, we use RAFT to propagate the object masks and provide a quantitative comparison with our method on the OVIS-Sparse dataset.
Specifically, we use flow between frames $t$ and $t-1$ provided by RAFT to propagate the predicted masks $\textbf{m}_{t-1}$ in the frame $t-1$ to frame $t$, and then calculate the mask IoUs between the propagated masks and the predicted masks to get the flow score. As the same with the motion score, the flow score is added to the original matching score of VIS methods.

We compare RAFT and our InstMove on the OVIS-Sparse dataset. Two SOTA VIS methods, \ie MinVIS and IDOL, are used.
The frames and annotations are kept every 1, 3, 5, or 7 frames (\ie Sparse-1/3/5/7) to simulate different FPS.
Note that RAFT is pretrained on a large number of datasets including FlyingChairs~\cite{dosovitskiy2015flownet}, FlyingThings~\cite{mayer2016large}, FlyingThings3D, Sintel~\cite{butler2012naturalistic}, KITTI-2015~\cite{menze2015object}, and HD1K~\cite{kondermann2016hci}, while the VIS datasets are relatively small, we train our motion model on the OVIS-Sparse training set that only contains 485 videos.
As shown in Table~\ref{table:RAFTonVIS}, our method outperforms the optical flow-based method in different FPS, which demonstrates the robustness and effectiveness of \OURS.

\begin{table}
\renewcommand\arraystretch{0.9}
    \small 
    \centering
    \resizebox{1\columnwidth}{!}{
    \begin{tabular}{l*5{c}}
    \toprule
       &Sparse-1   &Sparse-3 &Sparse-5  &Sparse-7  \\
    \midrule
    MinVIS~\cite{MinVIS}  &19.2 &18.9 &15.3 &15.1  \\
    MinVIS + RAFT &20.4 &19.6 &18.1 &16.3   \\
    MinVIS + InstMove &\textbf{20.8} &\textbf{20.0} &\textbf{18.2} & \textbf{16.7}    \\
    \midrule
    IDOL~\cite{IDOL}  &24.4 &21.3 &16.5 &14.1  \\
    IDOL + RAFT &25.7 &\textbf{21.5} &17.5 &15.2   \\
    IDOL + InstMove  &\textbf{27.0} &\textbf{21.5} &\textbf{18.8} & \textbf{16.2}    \\
    \bottomrule
    \end{tabular}}
    \caption{
    \textbf{Effects of instance-level motion module (\OURS) and pixel-level motion module (RAFT) on VIS task.}
    We report the mAP on the OVIS-Sparse validation set. \OURS outperforms the optical flow-based method in different FPS, which demonstrates the robustness and effectiveness of \OURS. Note that RAFT is pretrained on a large number of datasets while \OURS is only trained on 485 videos.
    }
    \label{table:RAFTonVIS}
    \vspace{-1ex}
\end{table}

\end{document}